\documentclass[letterpaper]{article} 
\usepackage{aaai25}  
\usepackage{times}  
\usepackage{helvet}  
\usepackage{courier}  
\usepackage[hyphens]{url}  
\usepackage{graphicx} 
\usepackage{booktabs}
\usepackage{natbib}  
\usepackage{caption} 
\nocopyright
\usepackage{algorithm}
\usepackage{algorithmic}
 \usepackage{amsmath}
\usepackage{amssymb} 
\usepackage{multirow}
\usepackage{tabularx}
\usepackage{ragged2e}
\usepackage{bm}
\usepackage{multirow}
\usepackage{tabularx}
\usepackage{ragged2e}
\usepackage[table]{xcolor}
\usepackage{newfloat}
\usepackage{listings}
\DeclareCaptionStyle{ruled}{labelfont=normalfont,labelsep=colon,strut=off} 
\floatstyle{ruled}
\newfloat{listing}{tb}{lst}{}
\floatname{listing}{Listing}

\title{UGOD: Uncertainty-Guided Differentiable Opacity and Soft Dropout for Enhanced Sparse-View 3DGS}

\author {
    Zhihao Guo\textsuperscript{\rm 1},
   Peng Wang\textsuperscript{\rm 1}\thanks{Corresponding author},
    Zidong Chen\textsuperscript{\rm 2},
    Xiangyu Kong\textsuperscript{\rm 3},
    Yan Lyu\textsuperscript{\rm 4},\\
    Guanyu Gao\textsuperscript{\rm 5},
    Liangxiu Han\textsuperscript{\rm 1}
}
\affiliations {
    \textsuperscript{\rm 1}Manchester Metropolitan University \quad
    \textsuperscript{\rm 2}Imperial College London \quad
    \textsuperscript{\rm 3}University of Exeter\quad \\
    \textsuperscript{\rm 4}Southeast University \quad
    \textsuperscript{\rm 5}Nanjing University of Science and Technology 
 
}

\usepackage{bibentry}

\begin{document}

\maketitle

\begin{abstract}
3D Gaussian Splatting (3DGS) has become a competitive approach for novel view synthesis (NVS) due to its advanced rendering efficiency through 3D Gaussian projection and blending. However, Gaussians are treated equally weighted for rendering in most 3DGS methods, making them prone to overfitting, which is particularly the case in sparse-view scenarios. To address this, we investigate how adaptive weighting of Gaussians affects rendering quality, which is characterised by learned uncertainties proposed. This learned uncertainty serves two key purposes: first, it guides the differentiable update of Gaussian opacity while preserving the 3DGS pipeline integrity; second, the uncertainty undergoes soft differentiable dropout regularisation, which strategically transforms the original uncertainty into continuous drop probabilities that govern the final Gaussian projection and blending process for rendering. Extensive experimental results over widely adopted datasets demonstrate that our method outperforms rivals in sparse-view 3D synthesis, achieving higher quality reconstruction with fewer Gaussians in most datasets compared to existing sparse-view approaches, e.g., compared to DropGaussian, our method achieves 3.27\% PSNR improvements on the MipNeRF 360 dataset.


\end{abstract}

\section{Introduction}
\label{introduction}


\begin{figure*}[h]
    \centering
	\includegraphics[width=1\textwidth]{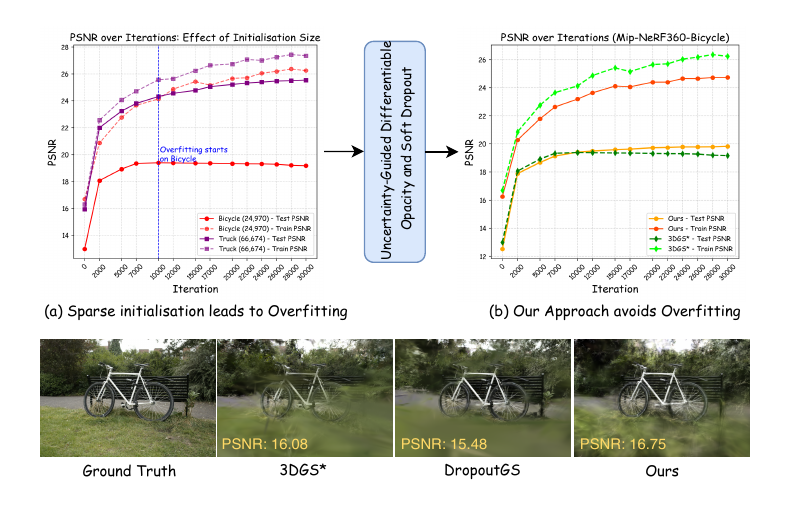}
	\caption{Peak Signal-to-Noise Ratio (PSNR) over iterations on the Mip-NeRF 360 Bicycle scene (24,970 initial Gaussians) and Tanks and Temples Truck scene (66,674 initial Gaussians). (a) Sparse initialisation leads to rapid overfitting and stagnant testing PSNR in baseline 3DGS*, while denser initialisation enables continued improvement. (b) The proposed work effectively suppress overfitting and improve generalisation, even under sparse initialisation. The bottom qualitative comparisons show that our approach yields better visual quality and higher PSNR than rivals.} 
    \label{fig:psnr-overfitting}
\end{figure*}


From 2D images as input to generate continuous high fidelity 3D reconstructed environments, which we also call it Novel View Synthetic (NVS), can contribute greatly to techniques like Digital Twinning, AR/VR and robotic embodiment~\cite{wang2024deep,fei20243d,xiong2024event3dgs,wang2024robot}. 
3D Gaussian Splatting (3DGS)~\cite{kerbl20233d} has emerged as a competitive approach for achieving real-time yet high-fidelity 3D scene synthesis. This is mostly attributed to its creative idea of representing scenes using a set of 3D Gaussians (ellipsoids) with learnable attributes like positions and colours. Once trained, these Gaussians are fed to the rasterisation-based forward rendering pipeline to be blended to yield the reconstructed scenes. Compared to classical rivals like Neural Radiance Fields (NeRF) and its variants~\cite{mildenhall2021nerf,guo2024depth} that rely on inefficient ray sampling, 3DGS enables efficient parallel rendering, making it a compelling choice for modern NVS applications. 

However, despite its advantages, 3DGS faces challenges in handling sparse-view scenarios, where the initialisation of Gaussians from Structure-from-Motion (SfM) point clouds is often incomplete or noisy. This leads to overfitting during optimisation, where the model fits training views excessively while failing to generalise to novel (testing) views. SfM-free methods such as DUSt3R~\cite{wang2024dust3r} and MASt3R~\cite{duisterhof2024mast3r} regress 3D points directly from images using pretrained models for Gaussian initialisation, which also suffer from overfitting, etc. As shown in Figure~\ref{fig:psnr-overfitting}, the baseline 3DGS* quickly stagnates and even declines in test PSNR due to overfitting, especially when initialised with sparse point clouds. 

Another under-explored question in existing 3DGS methods is: whether all Gaussians should be treated equally during rendering once training is complete~\cite{han2025view,li2024variational}. In practice, Gaussians exhibit strong view-dependence: a Gaussian may appear certain from one direction but highly uncertain from another. This issue is further exacerbated in sparse-view scenarios, where the initialisation of Gaussians is often incomplete or noisy. These observations motivate a systematic investigation into how adaptive, view-dependent treatment of Gaussians impacts rendering quality.

This paper aims to address these challenges by primarily investigating how reliable each Gaussian is for rendering, which is learned and characterised by their `uncertainty' using a neural network module that is integrated to the 3DGS pipeline. The uncertainty refers to a per-Gaussian, view-dependent parameter, learned by the neural network conditioned on spatial features (position, rotation, scale) and the current viewing direction: a higher uncertainty indicates that the Gaussian is less reliable and thus should be handled with propotional caution, while a lower uncertainty suggests greater confidence and indicates potential of further exploitation. Our approach ensures that uncertainty is learned in a fully differentiable manner, enabling two key functionalities: (1) adaptive modulation of opacity without disrupting the integrity of the 3DGS pipeline, making our method easily reusable by the research community; (2) application of soft, differentiable dropout regularisation, which transforms the predicted uncertainty into continuous drop probabilities that govern the final Gaussian projection and blending process. 
In summary, \textbf{our contributions} are as follows:


\begin{itemize}
    \item Gaussians Uncertainty Learning: We propose to learn an uncertainty for each Gaussian based on its spatial properties (position, scale, rotation) and the current viewing direction. This uncertainty captures per-Gaussian ambiguity within a view-dependent context, aligning with the fact that perceptions are view-dependent.
    \item Uncertainty-Guided Opacity Modulation: We utilise the learned uncertainty to modulate the opacity, which plays a critical role in rendering quality. In comparison, the opacity in literature has been treated as fixed when learned, which overlooks view-dependent unceratainty of Gaussians undermine the rendering quality. 
    \item Uncertainty-Guided Differentiable Soft Dropout: We introduce an uncertainty-guided soft dropout module guided by the learned uncertainty, which drops Gaussians with high uncertainty softly to further improve rendering quality. Together our work suppresses overfitting and improves rendering quality in sparse-view 3D reconstruction. 
\end{itemize}

\section{Related works}

\subsection{3D Gaussian Splatting}
3DGS has rapidly become a leading approach for NVS due to the tradeoff between rendering efficiency and quality. 
By representing scenes as collections of 3D Gaussians, rendering can be achieved by simply projects them onto the 2D image plane. This process, combined with depth sorting and $\alpha$-blending, enables efficient and high-fidelity scene reconstruction in real time. Recent works have further improved 3DGS by addressing camera pose sensitivity~\cite{yu2024mip} and refining point management for enhanced rendering quality~\cite{yang2024gaussian,zhang2024pixel,bulo2024revising}.

However, opacity modeling in 3DGS remains relatively underexplored, despite its critical role in accurate scene reconstruction and generalisation. Most existing methods treat opacity as a fixed or independently optimised parameter, overlooking its geometric and view-dependent nature. Only a handful of studies have investigated opacity optimization for 3DGS: Celarek~\cite{celarek2025does} provides a mathematical analysis of opacity-based versus extinction-based formulations in 3DGS and volumetric rendering. Talegaonkar~\cite{talegaonkar2024volumetrically} proposes volumetrically consistent 3D Gaussian rasterisation, which improves opacity computation by integrating 1D Gaussian densities along the ray. OMG~\cite{yong2025omg} introduces material-aware opacity modeling, linking opacity to material properties such as albedo and roughness. Nevertheless, these approaches do not explicitly model the geometric or view-dependent factors that influence opacity, leaving a gap in fully leveraging opacity for robust and generalisable 3DGS rendering.

\subsection{Sparse 3DGS Reconstruction}
\label{Sparse 3D Reconstruction}


While 3DGS has achieved remarkable rendering quality, its performance is highly dependent on dense input views for reliable Gaussian initialisation. Practically, dense view can be challenging, resulting in incomplete or noisy initialisation of Gaussians, which lead to model overfitting and undermines the performance of 3DGS in sparse-vew scenarios.

To address these challenges, various methods have been proposed for sparse-view 3DGS. DropGaussian~\cite{park2025dropgaussian} combats overfitting by selectively dropping low-contributing Gaussians, allowing the remaining ones to receive stronger gradients and contribute more effectively to optimization. CoR-GS~\cite{zhang2024cor} introduces point disagreement and rendering disagreement to quantify geometric and appearance inconsistencies between reconstructions. They find these metrics negatively correlate with reconstruction quality, making them useful for quality assessment. To reduce these disagreements, they introduce co-pruning for geometry refinement and pseudo-view co-regularization for appearance consistence. Although these approaches improve rendering under sparse supervision, they also introduce new limitations: some rely on external priors such as pretrained depth or diffusion models, increasing system complexity and computational requirements; others employ heuristic or non-differentiable dropout strategies, which restrict end-to-end training and adaptability to view-dependent uncertainty.

This work addresses sparse-view 3DGS from a new perspective by introducing Uncertainty-Guided Differentiable Soft Dropout directly into the splatting pipeline. Our approach is fully end-to-end trainable and does not require semantic labels or external pretrained models. The core idea is that not all Gaussians contribute equally to rendering quality, especially when initialised from sparse or noisy inputs, and their influence should be adaptively modulated based on view-dependent uncertainty.


\section{Methodology}
\subsection{Preliminaries}
\textbf{3D Gaussian Splatting
} 3DGS normally initialises Gaussians using point clouds like those from SfM, and Gaussians are defined by parameters like positions, rotations, scales, covariances, opacities, and colours. Representing the position of a Gaussian as $\mathbf{x}\in \mathbb{R}^{3 \times 1}$, we define a 3D Gaussian $G$ as follows:
\begin{equation}
G(\mathbf{x})=e^{-\frac{1}{2}(\mathbf{x}-\boldsymbol{\mu})^T \boldsymbol{\Sigma}^{-1}(\mathbf{x}-\boldsymbol{\mu})},
\end{equation}
where $\boldsymbol{\mu} \in \mathbb{R}^{3 \times 1}$ is the mean position, $\boldsymbol{\Sigma} \in \mathbb{R}^{3 \times 3}$ is the covariance matrix that is positive semi-definite, which can be factored as $\mathbf{\Sigma} = \mathbf{R} \mathbf{S} \mathbf{S}^T \mathbf{R}^T$, where $\mathbf{R} \in \mathbb{R}^{3 \times 3}$ is an orthogonal rotation matrix, and $\mathbf{S} \in \mathbb{R}^{3 \times 3}$ is a diagonal scale matrix. 3D Gaussians will be projected to 2D image space using the splatting-based rasterisation technique~\cite{zwicker2001surface}. Specifically, the transformation for projecting a 3D Gaussian onto the 2D image plane is approximated using a first-order Taylor expansion of the projection function at the Gaussian’s mean, expressed in the camera coordinate frame. The 2D covariance matrix $\boldsymbol{\Sigma}^{\prime}$, which describes the elliptical shape of each Gaussian in the image space, is then computed as: 
\begin{equation}
    \boldsymbol{\Sigma}^{\prime}=\mathbf{J W } \boldsymbol{\Sigma} \mathbf{W}^T \mathbf{J}^T, 
\end{equation}
where $\mathbf{J}$ is the Jacobian of the affine approximation of the projective
transformation, and $\mathbf{W}$ denotes the view transformation matrix. The colour of each pixel is calculated by blending sorted Gaussians based on the opacity $\alpha$:
\begin{equation}
c=\sum_{i=1}^n c_i \alpha_i \prod_{j=1}^{i-1}\left(1-\alpha_j\right),
\end{equation}
where $n$ is the number of points, $c_i$ is the color of the $i-th$ point, $\alpha_i$ can be obtained by evaluating a projected 2D Gaussian with covariance $\boldsymbol{\Sigma}^{\prime}$ multiplied with a learned opacity for each point.

\textbf{Opacity in 3DGS} In vanilla 3DGS, opacity $\alpha$ is treated as an independent parameter, optimised alongside other Gaussian parameters like positions and colours. Once trained, all Gaussian opacities will be equally used for rendering. In addition, opacity plays a key role in regulating the density of 3D Gaussians during training, to make sure Gaussians with very low $\alpha$ values are pruned, whereas remaining Gaussians are adjusted to refine the scene reconstruction. 

While this opacity-driven density control enhances both the efficiency and quality of the final rendered output, the opacity is not explicitly modelled in 3DGS. This overlooks the fact that not all Gaussians contribute equally to rendering quality, especially when initialised from sparse or noisy inputs. We will address this limitation by introducing uncertainty modelling into the 3DGS pipeline, which will be discussed in the next section.



\begin{figure*}[h]
    \centering
	\includegraphics[width=1\textwidth]{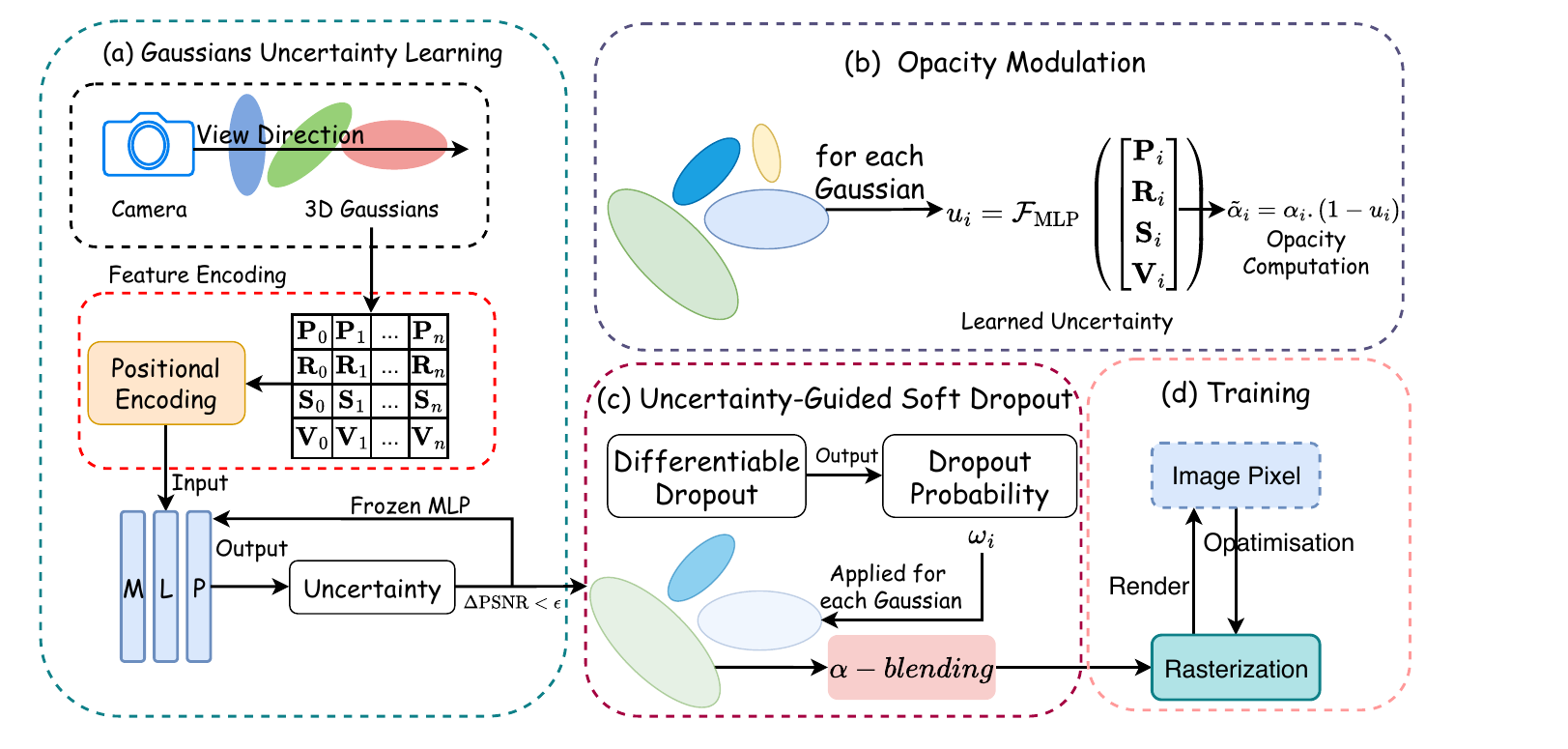}
	\caption{Overview of our proposed framework: Uncertainty-Guided Differentiable Opacity Modulation and Soft Dropout: (a) Gaussians Uncertainty Learning: we predict per-Gaussian view-dependent uncertainty using a proposed neural network conditioned on view direction and spatial features. 
(b) Uncertainty-Guided Opacity Modulation: The predicted uncertainty $u_i$ is used to modulate the learned opacity $\alpha_i$ via $\tilde{\alpha}_i = \alpha_i \cdot (1-u_i)$, reducing the contribution of ambiguous Gaussians during rendering. 
(c) Uncertainty-Guided Soft Dropout: To further suppress overfitting, we apply a differentiable soft dropout mechanism, using uncertainty to govern the soft dropout of each Gaussian. 
We also freeze the uncertainty learning neural network after overfitting is detected to ensure stability.} 
    \label{fig:overview}
\end{figure*}

\subsection{Gaussians Uncertainty Learning}
\label{Directional Uncertainty Prediction}

\textbf{Uncertainty Learning Module} We introduce and integrate a neural network into the 3DGS pipeline, which takes as input spatial features (position, rotation, scale) and the current viewing direction, and outputs a per-Gaussian uncertainty. This uncertainty is defined as view-dependent to capture the reliability of each Gaussian for rendering from a specific viewpoint. Essentially, the Gaussian uncertainty learning process can be formulated as 
\begin{equation}
    \mathcal{F}_{\text{MLP}}(\mathbf{I}_i ; \Theta)=\mathbf{u},
\end{equation}
where $\mathcal{F}_{\text{MLP}}$ is the neural network parameterised by $\Theta$, which is optimised during training by minimising the difference between the rendered images and
the reference images along with other parameters, $\mathbf{u}$ is a vector with entry $\quad u_i \in (0,1)$ represents the uncertainty of the $i-$th Gaussian $G_i$, and $\mathbf{I}_i$ wraps up neural network inputs and takes the form 
\begin{equation}\label{eq:original_input}
    \mathbf{I}_i = \big[\mathbf{P}^T_i, \mathbf{V}^T_i, \mathbf{R}^T_i, \mathbf{S}^T_i\big]^T \in \mathbb{R}^{13\times 1},
\end{equation}
which concatantes position $\mathbf{P}_i \in \mathbb{R}^{3\times1}$, view direction as a unit vector $\mathbf{V}_i \in \mathbb{R}^{3\times1}$ from the camera centre to $\mathbf{P}_i$, the scaling factor $\mathbf{S}_i \in \mathbb{R}^{3\times1}$, and the rotation represented by a quaternion $\mathbf{R}_i \in \mathbb{R}^{4\times1}$. 

It is worth noting that the low dimensionality of $\mathbf{I}_i$ fundamentally limits the neural network's representational capacity, constraining its ability to learn complex uncertainty patterns. This limitation is exacerbated by the spectral bias of neural networks~\cite{rahaman2019spectral}, which favor low-frequency functions and struggle with high-frequency scene details. 

\textbf{Input Ecoding} To overcome these constraints, we employ multilevel HashGrid encoding~\cite{muller2022instant} to transform $\mathbf{I}_i$ into a high-dimensional feature space. This dimensional expansion enables discrimination of subtle spatial variations that are indistinguishable in low-dimensional space. The high-dimensional representation makes complex patterns more linearly separable, allowing the neural network to capture both coarse and fine-scale details simultaneously. While one can encode the whole $\mathbf{I}_i$ using a single HashGrid, we find that encoding the position $\mathbf{P}_i$ separately yields better performance. We attribute this to the fact that positions of Gaussians tend to change smoothly (low frequency), and encode them to be high frequency can help better capture uncertainty. 



To be specific, the position $\mathbf{P}_i$ of Gaussian $G_i$ is encoded using a multilevel HashGrid encoder $H(\cdot)$ with $L$ levels ($L=6$ in our implementation) and $F=4$ features per level, yielding a 24-dimensional embedding. With the position encoding, Equation (\ref{eq:original_input}) can be rewritten as:

\begin{equation}\label{eq:encoded_input}
    \mathbf{I}_i = \big[H(\mathbf{P})^T_i, \mathbf{V}^T_i, \mathbf{R}^T_i, \mathbf{S}^T_i\big]^T \in \mathbb{R}^{34\times 1},
\end{equation}
with 
\begin{equation}
\label{eq6}
H(\mathbf{P}_i) = \oplus_{l=1}^{L} \mathrm{Interp}\big(T_l, \phi_l(\mathbf{P}_i \cdot r_l)\big) \in \mathbb{R}^{24\times 1},
\end{equation}
where $\oplus$ denotes concatenation, $T_l$ is the learnable feature table at level $l$, $\phi_l(\cdot)$ is a spatial hash function mapping scaled 3D coordinates to table indices, and $\mathrm{Interp}(\cdot)$ performs trilinear interpolation over lattice vertices. The resolution $r_l$ at each level increases geometrically as $r_l = r_{\text{base}} \cdot b^{l-1}$, where $b > 1$ is the per-level scaling factor. This encoding captures both low and high frequency spatial details efficiently.

\subsection{Uncertainty-Guided Differentiable Opacity Modulation and Soft Dropout}
\label{Uncertainty-Guided Opacity Masking and Dropout}
Our movitation of modelling Gaussian uncertainty is to suppress overfitting and improve the rendering quality of 3DGS, especially under sparse-view conditions. To achieve this, we propose two key mechanisms: Opacity Modulation and Uncertainty-Guided Soft Dropout. These mechanisms leverage the learned uncertainty to adaptively control Gaussian contributions during rendering, thereby enhancing robustness and generalisation.

\textbf{Opacity Modulation} The learned uncertainty $u_i \in (0, 1)$ of Gaussian $G_i$ is used to modulate the opacity $\alpha_i$ before rendering. Specifically, we compute the updated opacity of $G_i$ as:
\begin{equation}
\tilde{\alpha}_i=\alpha_i \cdot (1-u_i).
\end{equation}
This formulation acts as a soft gating mechanism, to ensure Gaussians with high uncertainty (i.e., low confidence from the current view direction) won't contribute as significantly to the rendered outputs as Gaussian with low uncertainty. By modulating the opacity according to the learned uncertainty, we suppress the influence of unreliable Gaussians.

\textbf{Uncertainty-Guided Soft Dropout} To further regularise learning and reduce overfitting, we propose a differentiable soft dropout that is also governed by uncertainty. Inspired by the concrete distribution~\cite{gal2017concrete}, we generate a continuous dropout probability $\omega_i \in (0,1)$ per Gaussian $G_i$ via:
\begin{equation}
\label{equation1}
\omega_i=1-\text{sigmoid}\Big(\frac{1}{\tau}\cdot\big(\log \frac{u_i}{1-u_i}+\log \frac{q_i}{1-q_i}\big)\Big),
\end{equation}
where $q_i \sim \mathcal{U}(0,1)$ is a random variable sampled from a uniform distribution for each Gaussian, which introduces controlled randomness in the dropout process, and $\tau > 0$ is a temperature hyperparameter (we use $\tau=0.1$). This produces a smooth, stochastic dropout probability $\omega_i$ that softly `drops' unreliable Gaussians while retaining gradient flow. 

\begin{figure}[t]
    \centering
	\includegraphics[width=0.45\textwidth]{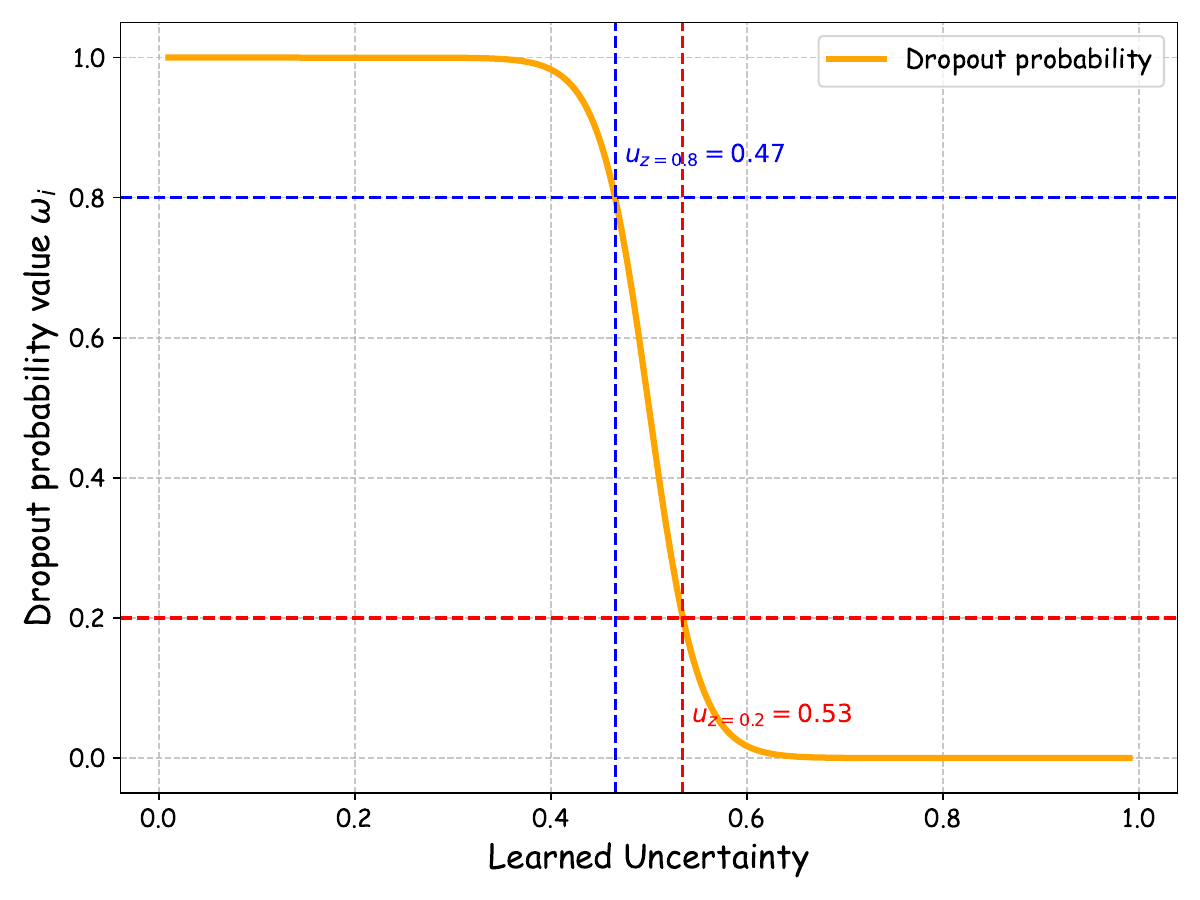}
	\caption{
Soft dropout probability $\omega$ as a function of the learned uncertainty $u$, with temperature $\tau=0.1$ and clamping range $[\omega_{\min}, \omega_{\max}] = [0.2, 0.8]$. When $u_i \approx 0.5$, Gaussians are most ambiguous and softly suppressed to reduce overfitting. For $u$ outside $[0.47, 0.53]$, clamping keeps $\omega$ stable, preserving gradient flow and allowing informative Gaussians to continuously contribute.
}

    \label{fig:sd}
\end{figure}

To ensure numerical stability and prevent collapse, we clamp the dropout probability $\omega_i$ following 
\begin{equation}\label{eq:clamp}
\tilde{\omega}_i = \text{clamp}(\omega_i, \, \omega_{\min}, \omega_{\max}),
\end{equation} 
with $[\omega_{\min}, \omega_{\max}]$ the range to clamp $\omega_i$ to. This clamping avoids extreme dropout values (too high or too low) that could lead to vanishing gradients or excessive suppression of potentially informative Gaussians. The final effective opacity $\bar{\alpha}_i$ for Gaussian $G_i$ is then computed as:
\begin{equation}\label{eq:final_opacity}
\bar{\alpha}_i = \tilde{\alpha}_i \cdot \tilde{\omega}_i.
\end{equation}


We set $\omega_{\min} = 0.2$ and $\omega_{\max} = 0.8$ in our experiments, as shown in Fig.~\ref{fig:sd}. This clamping ensures that the dropout probability $\omega_i$ remains within a stable range, corresponding to uncertainty values $u_i \in [0.47, 0.53]$. The insights behind are: when the learned uncertainty of a Gaussian is around 0.5, it is most ambiguous, i.e., neither clearly reliable nor unreliable, and is thus softly suppressed (partially dropped) to reduce its impact and mitigate overfitting. For Gaussians with uncertainty beyond $[0.47, 0.53]$, i.e., $\omega_i$ approaches the higher (1) or lower (0) bounds, we use the clamping machnisms shown in Equation (\ref{eq:clamp}) to prevent the probability from becoming too big or small, maintaining gradient flow and allowing potentially informative Gaussians to continue contributing to learning.

\subsection{3DGS Training Loss}
To train our uncertainty-guided 3DGS framework, we use the final effective opacity $\bar{\alpha}_i$ (Eq.~\ref{eq:final_opacity}) for rendering novel views $\hat{I}$, which are compared against ground-truth images $I$. Our objective is to jointly optimise for pixel-level accuracy and perceptual quality. Specifically, we employ a composite colour reconstruction loss that combines the mean absolute error (L1 loss) and a differentiable SSIM loss (D-SSIM) to encourage both sharpness and structural consistency:
\begin{equation}
\label{loss}
\mathcal{L}_{\text{colour}} = \mathcal{L}_1(\hat{I}, I) + \lambda \mathcal{L}_{\text{D-SSIM}}(\hat{I}, I),
\end{equation}
where $\mathcal{L}_1$ is the pixel-wise L1 loss, $\mathcal{L}_{\text{D-SSIM}}$ is the differentiable SSIM loss, and $\lambda$ is a balancing hyperparameter (set to 0.2 in our experiments).

To further mitigate overfitting, we monitor the PSNR improvement during training and freeze the uncertainty learning neural network once the improvement falls below a threshold $\epsilon$. This stabilises uncertainty estimation and prevents excessive adaptation to the training views.











\begin{algorithm}[t]
\caption{Uncertainty-Guided Differentiable Opacity Modulation and Soft Dropout}
\label{alg:uncertainty_dropout}
\begin{algorithmic}[1]
\REQUIRE 3D Gaussians $\{G_i\}$, each with position $\mathbf{P}_i$, view direction $\mathbf{V}_i$, scale $\mathbf{S}_i$, rotation $\mathbf{R}_i$, and opacity $\alpha_i$
\FOR{each Gaussian $G_i$}
    \STATE Encode $\mathbf{P}_i$ using multilevel HashGrid (Eq.~\ref{eq6})
    \STATE Form feature vector $\mathbf{I}_i = [H(\mathbf{P}_i)^T, \mathbf{V}_i^T, \mathbf{R}_i^T, \mathbf{S}_i^T]^T$
    \STATE Predict uncertainty $u_i = \mathcal{F}_{\text{MLP}}(\mathbf{I}_i; \Theta)$
    \IF{PSNR improvement $\Delta \text{PSNR} < \epsilon$}
        \STATE Freeze MLP parameters $\Theta$ to stabilise uncertainty learning
    \ENDIF
    \STATE Compute modulated opacity $\tilde{\alpha}_i = \alpha_i \cdot (1 - u_i)$
    \STATE Sample $q_i \sim \mathcal{U}(0,1)$
    \STATE Compute soft dropout probability $\omega_i$ via Concrete distribution (Eq.~\ref{equation1})
    \STATE Clamp mask: $\tilde{\omega}_i = \text{clamp}(\omega_i, \omega_{\min}, \omega_{\max})$
    \STATE Final opacity: $\bar{\alpha}_i = \tilde{\alpha}_i \cdot \tilde{\omega}_i$
\ENDFOR
\STATE Render image $\hat{I}$ using $\{\bar{\alpha}_i\}$ and Gaussian parameters
\STATE Compute colour loss $\mathcal{L}_{\text{colour}}$ (Eq.~\ref{loss})
\end{algorithmic}
\end{algorithm}

\section{Experiments}
\label{Experiments}
\subsection{Dataset and Implementation}
\textbf{Dataset} We test our methods on widely adopted datasets for 3DGS, including: 1) MiPNeRF 360~\cite{barron2022mip} that provides real world scenes. This dataset better reflects real world small-scale reconstruction tasks. We take 24 views (aligning with sparse-view setting) for training and the rest for testing. 2) MVimgNet~\cite{yu2023mvimgnet}, which is a large-scale dataset of multi-view images. We also use 24 input views for training and others for testing.

\textbf{Implementation} All experiments are conducted using the PyTorch
framework on NVIDIA RTX A100. Our modules are integrated with the 3DGS* and is adapted for sparse-view reconstrubtion. The training iteration for our method and the benchmark methods is set to 6,000 following DropGaussian for fair comparison. We set the HashGrid encoding configuration to $(6, 0, 0, 0)$ for position, view direction, scale, and rotation respectively. The soft dropout temperature $\tau$ is set to 0.2 for Mip-NeRF 360 and 0.1 for MVImgNet. The uncertainty learning neural network is frozen when the PSNR improvement $\Delta \text{PSNR}$ falls below the threshold $\epsilon = 0.2$. Notably, we do not apply the opacity reset mechanism as used in original 3DGS-based methods because our uncertainty-guided opacity modulation and soft dropout already prevent opacity oversaturation and overfitting. These mechanisms dynamically downweight unreliable Gaussians during training, making periodic opacity resets unnecessary for convergence or stability.

\begin{table*}[thbp]
  \centering
  \footnotesize
  \begin{tabular}{lcccccccccccc}
    \toprule
    \multirow{2}{*}{Methods} 
    & \multicolumn{4}{c}{Mip-NeRF360} 
    & \multicolumn{4}{c}{MVImgNet} \\
    \cmidrule(lr){2-5} \cmidrule(lr){6-9}
    & PSNR$\uparrow$ & SSIM$\uparrow$ & LPIPS$\downarrow$ & Gaussians$\downarrow$
    & PSNR$\uparrow$ & SSIM$\uparrow$ & LPIPS$\downarrow$ & Gaussians$\downarrow$ \\
    \midrule
    3DGS*~\cite{kerbl3Dgaussians} 
    & 18.42 & \underline{0.56} & \textbf{0.44} & 1040939
    & 25.77 & 0.85 & 0.18 & 1536370 \\
    
    DropGaussian~\cite{park2025dropgaussian} 
    & 18.34 & 0.55 & 0.47 & \textbf{669839}
    & 25.35 & 0.83 & 0.22 & \textbf{972767} \\

    CoR-GS~\cite{zhang2024cor} 
    & \underline{18.72} &0.55  &\underline{0.45}  & \underline{865395}
    & 25.54 & 0.84 & 0.20 & \underline{1269568} \\
    
    \textbf{Ours}
    & \textbf{18.94} & \textbf{0.57} & \textbf{0.44} & 878194
    & \textbf{26.02} & \textbf{0.85} & \textbf{0.17} & 1467984 \\
    
    \bottomrule
  \end{tabular}
\caption{Comparison of baseline methods and our method on the Mip-NeRF360 and MVImgNet datasets. \textbf{Bold} denotes best, \underline{underline} denotes second-best. Our method consistently achieves the best sparse rendering quality while maintaining a compact Gaussian representation. More results can be found in Appendix.}
  \label{tab:merged_mip_mv}
\end{table*}

\textbf{Metrics} We compare our method against state-of-the-art sparse-view NVS approaches based on commonly used metrics: PSNR, Structural Similarity Index Measure (SSIM)~\cite{wang2004image}, and Learned Perceptual Image Patch Similarity (LPIPS)~\cite{zhang2018unreasonable} on the rendered images in the test views. Our method outperforms rivals on most scenes across all three metrics, demonstrating improvements in both pixel-level accuracy (PSNR) and perceptual quality (SSIM \& LPIPS). 
\subsection{Sparse-View Synthesis Results}
\label{Comparison with Sparse-View Scenes}

\textbf{Quantitative Results} The evaluation results of our method and the benchmarking methods on Mip-NeRF 360 and MVImgNet datasets are shown in Table~\ref{tab:merged_mip_mv}. We can see that our method achieves the best or second-best results in all scenes across all metrics. For example, as show in Appendix Table~\ref{tab:results_full} in the `bonsai' scene, we improve PSNR from 21.79 (3DGS), 22.01 (CoR-GS) and 21.79 (DropGaussian) to 22.27, and improve SSIM from 0.80 and 0.79 to 0.81, while using the fewest 3D Gaussians. Similarly, in the stump scene, we achieve the best PSNR (15.31) and SSIM (0.36), while maintaining a compact representation using the fewest 3D Gaussians. On the MVImgNet dataset (Appendix Table~\ref{tab:results_nv}), our method also achieves top performance in all scenes, where we consistently outperform both baselines in PSNR and LPIPS while maintaining efficient Gaussian numbers. These results confirm that our method achieves higher fidelity, better perceptual quality, and improved compactness across diverse and complex scenes.
\begin{figure*}[htbp]
    \centering
	\includegraphics[width=1\textwidth]{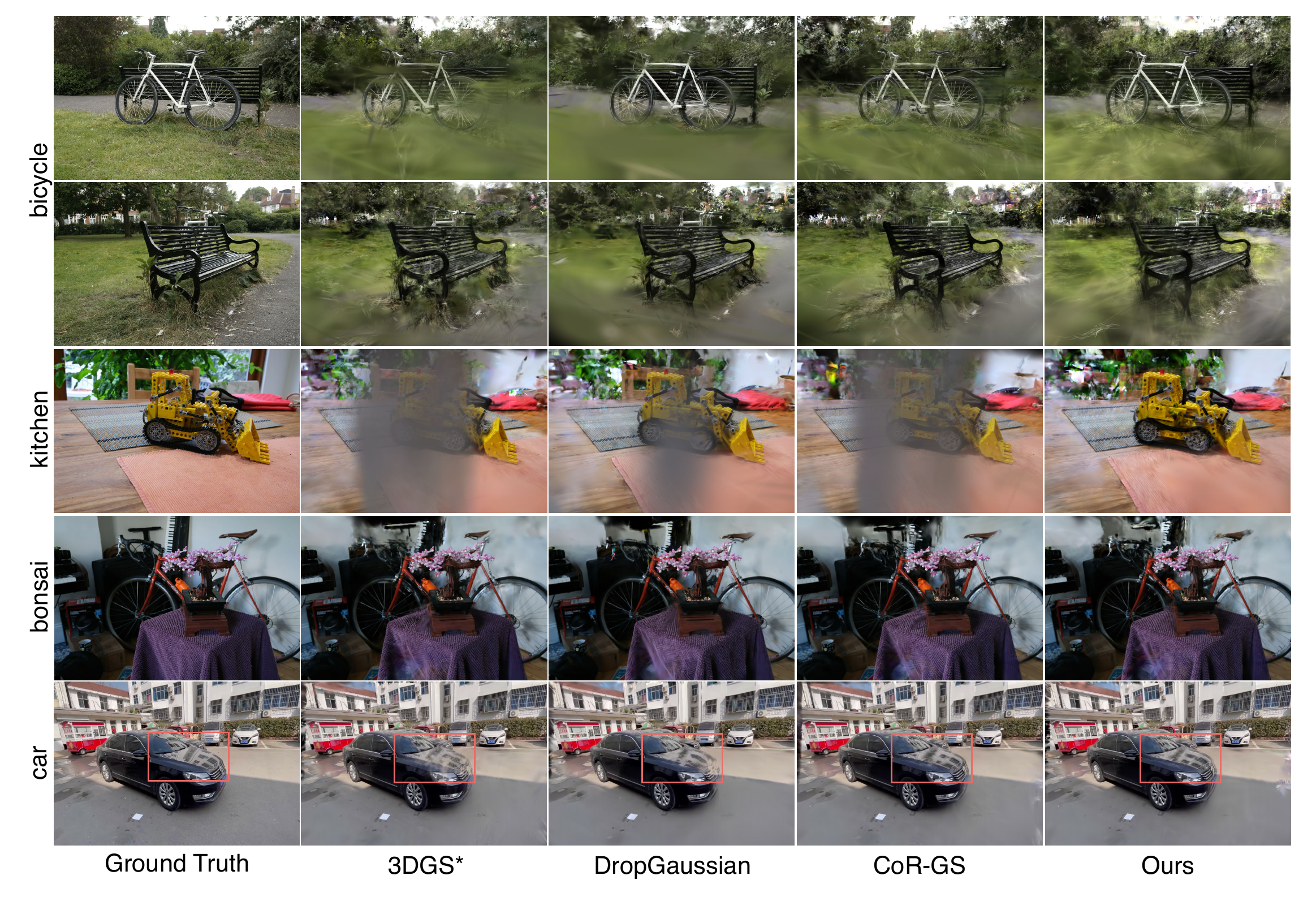}
	  \caption{Qualitative comparison of NVS on the Mip-NeRF 360 and MVImgNet datasets. 
  We compare our method with 3DGS*, DropGaussian and CoR-GS across multiple challenging scenes: bicycle, kitchen, and bonsai from MipNeRF 360, and car from MVImgNet. Our method produces more faithful geometry and preserves sharper structural details (e.g., the bicycle frame, excavator body, and flower arrangement) while mitigating artifacts and overfitting. 
  Notably, DropGaussian struggles with over-blurring in occluded regions (e.g., excavator centre), and 3DGS* suffers from noisy or incomplete opacity modelling. Our uncertainty-guided opacity and soft dropout allows better generalisation across both simple scenes (e.g., sparse backgrounds) and complex scenes with clutter, occlusion, or thin structures.
  }
    \label{fig:quality-mipnerf}
\end{figure*}

\textbf{Qualitative Results} 
As shown in Figure~\ref{fig:quality-mipnerf}, our method consistently achieves higher visual fidelity compared to 3DGS*, DropGaussian and CoR-GS across a range of scenes from the Mip-NeRF 360 dataset. For instance, in the `bicycle' scene, our method reconstructs sharper edges and cleaner silhouettes, while DropGaussian and CoR-GS tends to over-blur occluded or uncertain regions. As shown in Figure~\ref{fig:quality-mipnerf}, in the `kitchen' and `bonsai' scenes, our approach yields more complete structures and visually plausible textures without the opacity holes and noisy blending observed in other methods, and our method reconstructs fine-grained structures and reflective regions more faithfully than 3DGS*, DropGaussian, and CoR-GS. For example, in the `car' scene, the specular highlights and contours around the wind shield and hood are better preserved, while DropGaussian exhibits blur and distortion. These improvements demonstrate the effectiveness of our uncertainty-guided opacity modulation and soft dropout design.

\begin{figure*}[h]
    \centering
	\includegraphics[width=1\textwidth]{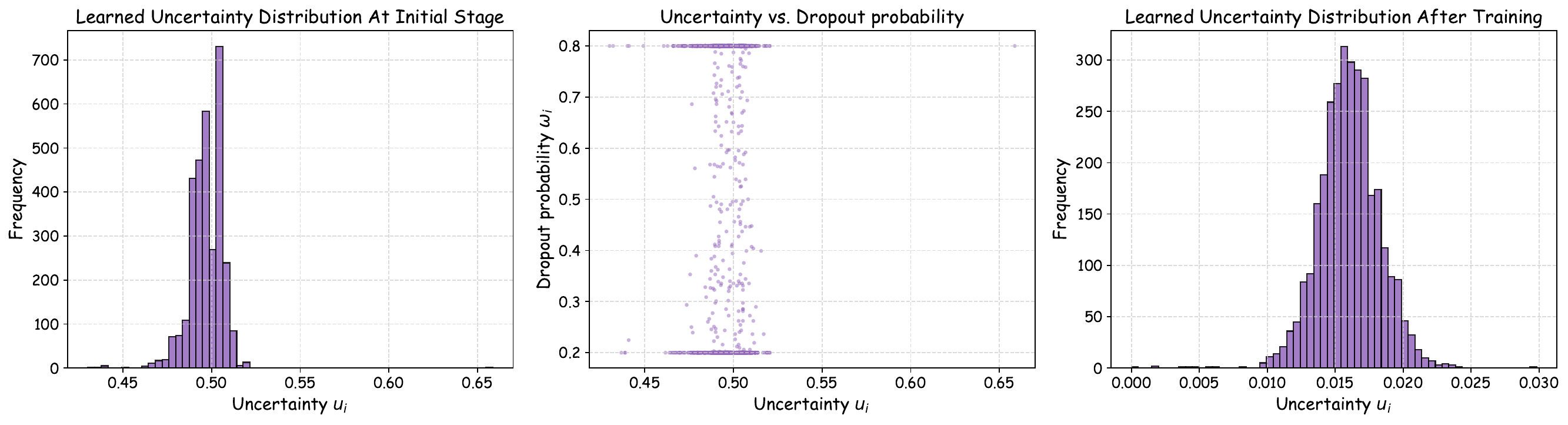}
	\caption{Illustration of uncertainty, soft dropout probability, and their relationships. From left to right, 1) the first figure shows the uncertainty histogram at the initial stage, where the majority of Gaussians are ambiguous (uncertainty around 0.5); 2) these Gaussians will be assigned to be either certain or uncertain by Equation (\ref{equation1}), and the rationale is given that Gaussians are ambiguous, one can randomly classify them as certain or uncertain as shown in the second figure; 3) The third figure shows the uncertainty distribution when training is complete, where the uncertainty of most Gaussians converges to 0, meaning we have got low uncertainty Gaussians that we can rely on for rendering. 
}\label{fig:un}
\end{figure*}
\subsection{Ablation Study}
Ablation studies are conducted to investigate the impact of HashGrid encoding configuration, Gaussians uncertainty on our method. We also test our method under dense views to show it can cope with dense view scenarios.

\textbf{HashGrid encoding configuration} We conduct an ablation study on the MipNeRF 360 `kitchen' scene under sparse view settings, as shown in Table~\ref{tab:hashgrid_ablation}. Each configuration is represented as $(\textbf{P}, \textbf{S}, \textbf{R}, \textbf{V})$, denoting the number of encoding dimensions allocated to position $\textbf{P}$, scale $\textbf{S}$, rotation $\textbf{R}$, and view direction $\textbf{V}$, respectively. We observe that encoding only the position (e.g., $(6,0,0,0)$) achieves the best rendering quality (PSNR: 19.15, SSIM: 0.66, LPIPS: 0.39), while adding scale, rotation, or view direction consistently degrades performance. This can be explained as HashGrid encoding is specifically designed to capture static, spatially local patterns across multiple scales, making it particularly effective for representing 3D positions. In contrast, view direction is a dynamic, per-ray input that lacks spatial coherence, encoding it using spatial HashGrids introduces aliasing and inconsistent gradients, leading to unstable training. Moreover, in 3DGS framework, both rotation and scaling are already explicitly modelled and directly applied during the anisotropic splatting process. Encoding them again is not only redundant but also potentially harmful, as it entangles global transformation parameters with spatial encodings, disrupting learning dynamics.


\begin{table}[tbp]
\centering
\footnotesize
\begin{tabular}{cccc}
\toprule
 (P, S, R, V) & PSNR $\uparrow$ & SSIM $\uparrow$ & LPIPS $\downarrow$ \\
\midrule
(5,0,0,0) & 18.88 & 0.64 & 0.41 \\
(5,0,0,1) & 18.50 & 0.64 & 0.41 \\
(6,0,0,0) & \textbf{19.15} & \textbf{0.66} & \textbf{0.39} \\
(6,1,1,0) & 18.96 & 0.65 & 0.40 \\
(7,0,0,0) & 18.69 & 0.63 & 0.43 \\
\bottomrule
\end{tabular}
\caption{HashGrid input encoding configurations for the MipNeRF 360 kitchen scene under 24 sparse input views. Each tuple represents the dimensional allocation for position, scaling, rotation, and view direction. \textbf{Bold} is \textbf{Best}}.
\label{tab:hashgrid_ablation}
\end{table}

\textbf{Gaussians Uncertainty Analyse}
While an ideal mapping from uncertainty $u_i$ to soft dropout probability $\omega_i$ follows a sigmoid-shaped function (Fig.~\ref{fig:sd}), we observe in practice that by introducing a temperature $\tau$ and some ramdomness through $q_i \sim \mathcal{U}(0,1)$ into Equation (\ref{equation1}), we can achieve better overall performance as indicated by Figure \ref{fig:un}. We attribute this to the fact that when the temperature $\tau$ is small (e.g., $\tau = 0.1$, the smaller $\tau$ is the steeper sigmoid function is), even minor fluctuations in $u_i$ can shift the resulting $\omega_i$ sharply toward either extreme after clamping. This is particularly helpful when the majority of Gaussians are ambiguous (i.e., $u_i \approx 0.5$), as it allows us to randomly classify them as certain or uncertain, which is beneficial for training.

\textbf{Dense Views Study} We also integrate our method to dense views to evaluate the NVS tasks. As show in Table~\ref{tab:bicycle_selected_iters}, we present the comparison on the `bicycle' scene between our method and 3DGS* at selected training iterations. Our method demonstrates a consistent improvement in rendering quality throughout the training process, as reflected by the steadily increasing PSNR, SSIM and decreasing LPIPS values. In contrast, 3DGS* begins to show signs of overfitting after iteration 20,000, with slight degradation in PSNR and LPIPS. 

\begin{table}[tbp]
\centering
\footnotesize
\begin{tabular}{ccccccc}
\toprule
\multirow{2}{*}{Iter} & \multicolumn{3}{c}{Ours} & \multicolumn{3}{c}{3DGS*} \\
\cmidrule(lr){2-4} \cmidrule(lr){5-7}
 & PSNR & SSIM & LPIPS & PSNR & SSIM & LPIPS \\
\midrule
5,000  & 18.68 & 0.55 & 0.39 & \textbf{18.92} & \textbf{0.59} & \textbf{0.33} \\
10,000 & 19.42 & 0.59 & 0.33 & \textbf{19.39} & \textbf{0.63} & \textbf{0.28} \\
15,000 & \textbf{19.59} & 0.61 & 0.31 & 19.36 & \textbf{0.64} & \textbf{0.26} \\
20,000 & \textbf{19.73} & \textbf{0.65} & \textbf{0.24} & 19.33 & 0.64 & 0.25 \\
25,000 & \textbf{19.79} & \textbf{0.66} & \textbf{0.24} & 19.30 & 0.64 & 0.25 \\
30,000 & \textbf{19.83} & \textbf{0.66} & \textbf{0.23} & 19.28 & 0.64 & 0.24 \\
\bottomrule
\end{tabular}
\caption{Evaluation on the MipNeRF 360 bicycle scene at selected iterations for Ours and 3DGS*. \textbf{Bold} indicates \textbf{Best}. }
\label{tab:bicycle_selected_iters}
\end{table}

\section{Conclusion}
We investigate and provide a solution to an under-explored question in 3DGS: if we treat Gaussians differently (compared to treating them equally in conventionally 3DGS methods) once trained, how will that affect the rendering quality? The answer to this question is of particular interests to sparse-view 3DGS as a measure and exploitation of view-dependent uncertainty is promising to suppress overfitting and improve rendering quality. We answer the question by learning a view-dependent uncertainty for each Gaussian, conditioned on its spatial properties and viewing direction. This uncertainty is then exploited in two key ways: first, to adaptively modulate opacity in a differentiable manner, thereby reducing the impact of unreliable Gaussians; and second, to drive a soft, differentiable dropout mechanism that regularises training by encouraging the model to suppress Gaussians with ambiguous contributions. Experimental results demonstrate consistent improvements in sparse-view rendering quality and robustness.

\bibliography{aaai25}

\clearpage
\onecolumn
\section{Appendix}\label{sec:appendix}
In this section, we provide the full quantitative results for all evaluated scenes across the MipNeRF 360 and MVImgNet datasets. The results include PSNR, SSIM, LPIPS, and the total number of Gaussians used for reconstruction. Our method is compared against three recent baselines: 3DGS*, DropGaussian, and CoR-GS. We highlight the best and second-best results in bold and underline, respectively. These tables supplement the main paper by offering a comprehensive view of the per-scene performance, demonstrating that our approach consistently achieves a strong balance between sparse view rendering quality and Gaussian compactness.
\begin{table*}[h]

  \centering
  \footnotesize

  \begin{tabular}{lcccccccc}
    \toprule
    Method
    & \multicolumn{4}{c}{bicycle}
    & \multicolumn{4}{c}{kitchen} \\
    \cmidrule(lr){2-5} \cmidrule(lr){6-9}
    & PSNR$\uparrow$ & SSIM$\uparrow$ & LPIPS$\downarrow$ & Gaussians$\downarrow$
    & PSNR$\uparrow$ & SSIM$\uparrow$ & LPIPS$\downarrow$ & Gaussians$\downarrow$ \\
    \midrule
    3DGS*~\cite{kerbl3Dgaussians} & \underline{15.13} & 0.28 &\textbf{ 0.58} & 1322955 & 18.73 & 0.64 & \underline{0.41} & 663026 \\
    DropGaussian~\cite{park2025dropgaussian}                     & 14.90 & \textbf{0.31} & \underline{0.59} & \textbf{744891}  & 18.92     & \underline{0.63}    & 0.43    & \textbf{494612} \\
        CoR-GS~\cite{zhang2024cor}  & 15.72 & \underline{0.30} & \underline{0.59} & 1152403 &  \underline{19.03}    & \textbf{0.65} & \underline{0.41} & \underline{591237}\\
    Ours & \textbf{15.99} & \textbf{0.31} & \textbf{0.58} & \underline{1013827} & \textbf{19.17}     & \textbf{0.65}   & \textbf{0.40 }  & 635714 \\
    \midrule
    Method
    & \multicolumn{4}{c}{bonsai} 
    & \multicolumn{4}{c}{garden} \\
    \cmidrule(lr){2-5} \cmidrule(lr){6-9}
    & PSNR$\uparrow$ & SSIM$\uparrow$ & LPIPS$\downarrow$ & Gaussians$\downarrow$
    & PSNR$\uparrow$ & SSIM$\uparrow$ & LPIPS$\downarrow$ & Gaussians$\downarrow$ \\
    \midrule
    3DGS*~\cite{kerbl3Dgaussians} & 21.79& 0.80 & 0.31 &695974  & \underline{19.68} & \textbf{0.53} & \textbf{0.39} & 1731954\\
    DropGaussian~\cite{park2025dropgaussian}                     & 21.79    & 0.79    & 0.33   & 585740      & 19.48    & \underline{0.47}    & 0.47   & \textbf{1147084} \\
       CoR-GS~\cite{zhang2024cor}  & \underline{22.01} & \underline{0.80} &\textbf{ 0.30}  &\underline{584920}&  19.61    & \underline{0.47}  & \underline{0.41}  &\underline{1401520} \\
    Ours               & \textbf{22.27}    & \textbf{0.81 }   & \textbf{0.30 }  & \textbf{584832}       & \textbf{19.85}     & \textbf{0.53}    & \textbf{0.39 }  & 1615037 \\
    \midrule
    Method 
    & \multicolumn{4}{c}{counter} 
    & \multicolumn{4}{c}{stump} \\
    \cmidrule(lr){2-5} \cmidrule(lr){6-9}
    & PSNR$\uparrow$ & SSIM$\uparrow$ & LPIPS$\downarrow$ & Gaussians$\downarrow$
    & PSNR$\uparrow$ & SSIM$\uparrow$ & LPIPS$\downarrow$ & Gaussians$\downarrow$ \\
    \midrule
    3DGS*~\cite{kerbl3Dgaussians} & 19.96 & \underline{0.74} & \textbf{0.36} & 617813  & \underline{15.22} & \underline{0.35} & \textbf{0.57}& 1213913 \\
    DropGaussian~\cite{park2025dropgaussian}                     & 20.74     & 0.73    & 0.38   & \textbf{500889}       &   14.24  & \underline{0.35}   & 0.60    & \underline{916739} \\
        CoR-GS~\cite{zhang2024cor}  &\underline{20.80}  & 0.73 & \underline{0.37} & 523274 & 15.16     & 0.34  &0.59  &939018 \\
    Ours              & \textbf{21.06 }   & \textbf{0.75}   & \textbf{0.36}    & \underline{515422}       & \textbf{15.31 }    & \textbf{0.36}   & \underline{0.58}    & \textbf{904329} \\
    \bottomrule
  \end{tabular}
    \caption{Comparison of various methods on MipNeRF 360 dataset. Each scene contains 24 views. \textbf{Best} results are in \textbf{bold}, and \underline{second-best} results are underlined.}

  \label{tab:results_full}
\end{table*}

\begin{table*}[h]

  \centering
  \footnotesize
  
  \begin{tabular}{lcccccccc}
    \toprule
    Method
    & \multicolumn{4}{c}{bench}
    & \multicolumn{4}{c}{bicycle} \\
    \cmidrule(lr){2-5} \cmidrule(lr){6-9}
    & PSNR$\uparrow$ & SSIM$\uparrow$ & LPIPS$\downarrow$ & Gaussians$\downarrow$
    & PSNR$\uparrow$ & SSIM$\uparrow$ & LPIPS$\downarrow$ & Gaussians$\downarrow$ \\
    \midrule
    3DGS*~\cite{kerbl3Dgaussians} & \underline{25.79} & \textbf{0.85} & \textbf{0.15} & 2059825 & 24.59 & 0.85 & 0.15 & 3061480 \\
    DropGaussian~\cite{park2025dropgaussian}                     & 25.04  & \underline{0.82} & \underline{0.20} & \textbf{1257547}  & 23.70     & 0.82    & 0.20    & \textbf{1791593} \\
      CoR-GS~\cite{zhang2024cor}  & 25.42 & 0.84 & 0.18 & \underline{1658686} &    24.14  & 0.84  & 0.18 &\underline{2426536} \\
    Ours & \textbf{25.98} & \textbf{0.85} & \textbf{0.15} & 1930662 & \textbf{24.67}    & \textbf{0.86}  & \textbf{0.14} & 2818824 \\
    \midrule
    Method
    & \multicolumn{4}{c}{car} 
    & \multicolumn{4}{c}{chair} \\
    \cmidrule(lr){2-5} \cmidrule(lr){6-9}
    & PSNR$\uparrow$ & SSIM$\uparrow$ & LPIPS$\downarrow$ & Gaussians$\downarrow$
    & PSNR$\uparrow$ & SSIM$\uparrow$ & LPIPS$\downarrow$ & Gaussians$\downarrow$ \\
    \midrule
    3DGS*~\cite{kerbl3Dgaussians} & \underline{28.61}& \underline{0.92}& \textbf{0.16} &\underline{397803} & 27.49 & \textbf{0.87} & \underline{0.22} & 1160568\\
    DropGaussian~\cite{park2025dropgaussian}                     & 28.14    & 0.91   & 0.18  & \textbf{325706}     & \underline{27.65}   & 0.85   & 0.27    & \textbf{770096} \\
       CoR-GS~\cite{zhang2024cor}  &  28.38& 0.91 & \underline{0.17} &  431755&    27.57  & \underline{0.86}  &  0.24& \underline{965332}\\
    Ours               & \textbf{28.83}    & \textbf{0.93}   & \textbf{0.16}  & 426449      & \textbf{28.31}     &\textbf{0.87}   & \textbf{0.21}  & 1202907 \\
    \midrule
    Method 
    & \multicolumn{4}{c}{ladder} 
    & \multicolumn{4}{c}{suv} \\
    \cmidrule(lr){2-5} \cmidrule(lr){6-9}
    & PSNR$\uparrow$ & SSIM$\uparrow$ & LPIPS$\downarrow$ & Gaussians$\downarrow$
    & PSNR$\uparrow$ & SSIM$\uparrow$ & LPIPS$\downarrow$ & Gaussians$\downarrow$ \\
    \midrule
    3DGS*~\cite{kerbl3Dgaussians} & 20.38 & \textbf{0.68} & \underline{0.26} & 2202016  &\underline{27.78}  & \textbf{0.90} & \underline{0.14}& 336527 \\
    DropGaussian~\cite{park2025dropgaussian}                     & \underline{20.39}     & \underline{0.67}    & 0.29   & \textbf{1423981}     &   27.18  & \textbf{0.90}  &0.16   & \textbf{267679}\\
       CoR-GS~\cite{zhang2024cor}  &  20.35& \underline{0.67} & 0.28 & \underline{1812998} & 27.38     & \underline{0.89}  & 0.15  &322103 \\
  
    Ours              & \textbf{20.42}  & \textbf{0.68}  & \textbf{0.25}   & 2109951     & \textbf{27.90}  & \textbf{0.90}  & \textbf{0.13}    & \underline{319112}\\
    \bottomrule
  \end{tabular}
  \caption{Comparison of various methods on MVimgNet dataset. Each scene contains 24 views. \textbf{Best} results are in \textbf{bold}, and \underline{second-best} results are underlined.}

  \label{tab:results_nv}
\end{table*}

\end{document}